\documentclass[runningheads]{llncs}
\usepackage[T1]{fontenc}
\usepackage{graphicx}
\usepackage{xcolor}
\usepackage{colortbl}
\usepackage{hyperref}
\hypersetup{
    colorlinks=true,
    linkcolor=black,
    citecolor=black,
    urlcolor=blue
}
\usepackage{amsmath} 
\usepackage{float}
\usepackage{booktabs}
\usepackage{tabularx}

\usepackage{color}

\begin{document}
\title{Multi-Dimensional Evaluation of LLMs for Grammatical Error Correction\thanks{Accepted at AIED 2026}}

%
\author{Adnan Labib\inst{1}\orcidID{0009-0004-9916-6533} \and
Qiao Wang\inst{2}\orcidID{0000-0001-9273-7082} \and
Yixuan Huang\inst{3}\orcidID{0009-0001-8628-5128} \and
Zheng Yuan\inst{4}\orcidID{0000-0003-2406-1708}}

\authorrunning{A. Labib et al.}

\institute{King's College London, London, United Kingdom\\
\email{adnan.1.labib@kcl.ac.uk}\\
\and
Hosei University, Tokyo, Japan\\
\email{judy.wang@hosei.ac.jp}\\
\and
Waseda University, Tokyo, Japan\\
\email{yixuan.huang@moegi.waseda.jp}\\
\and
University of Sheffield, Sheffield, United Kingdom\\
\email{zheng.yuan1@sheffield.ac.uk}}

\maketitle

\setcounter{footnote}{0}

\begin{abstract}

Automated assistants for Grammatical Error Correction are now embedded in educational platforms serving millions of learners, yet three critical gaps remain in this domain: (1) latest-generation Large Language Models (LLMs) lack comprehensive evaluation on grammar correction tasks; (2) whether combining these LLMs improves correction quality is unexplored; and (3) the extent to which reference-based metrics underestimate GEC system performance has not been adequately quantified. In this study, first, we evaluate latest-generation LLMs on edit precision, fluency preservation, and meaning retention, showing fine-tuned GPT-4o achieves state-of-the-art performance across all three dimensions. Second, through grammatical error type analysis we demonstrate that individual LLMs exhibit highly similar error correction patterns ($\rho=0.947$). Third, we show that reference-based metrics underestimate GEC performance with 73.76\% of GPT-4o corrections different from gold standards being equally valid or even superior. These GEC evaluation findings equip educators with guidance for selecting GEC assistants that enhance rather than constrain student linguistic development. We make our data, code, and models publicly available.\footnote{\url{https://github.com/adnanlabib1509/evaluation-of-llm-for-gec}}

\end{abstract}
\section{Introduction}

Language learners, particularly in second language acquisition contexts, require consistent, accurate feedback on their writing to develop grammatical competence. However, providing detailed correction to every student is resource-intensive, particularly in large classrooms where teacher-student ratios make individualized feedback infeasible. Automated Grammatical Error Correction (GEC) systems offer a potential solution by serving as AI assistants which takes a student's erroneous sentence as input and provides feedback by automatically producing a grammatically correct version (of that sentence) in real-time.

While newer LLMs might be effective as GEC assistants, comprehensive evaluations of latest-generation models are notably absent as previous studies evaluated older LLMs like LLaMA-2 \cite{omelianchuk_2024}. Moreover, existing GEC research predominantly relies on single metric, typically ERRANT F$_{0.5}$ \cite{bryant_2017}, which evaluates correction precision but neglect dimensions like fluency preservation and meaning retention, both of which are vital feedback for language learning students.

Historically, ensembles outperformed individual systems by combining complementary strengths, but shared training objectives (next-token prediction), similar pre-training corpora and model designs (Transformers) suggest latest LLMs may exhibit similar performance individually.

Additionally, reference-based evaluation metrics of GEC systems penalize valid alternative corrections simply because they differ from provided references. This issue is particularly problematic with LLMs, which introduce stylistic enhancements that frequently diverge from gold standards while remaining grammatically valid \cite{Bryant_2023}. While reference-free metrics like IMPARA \cite{Maeda_2022} attempt to assess grammatical acceptability without references, they exhibit bias toward minimal edits \cite{Bryant_2023}. These evaluation limitations create substantial uncertainty about true GEC system performance, particularly in high-stakes educational settings where correction quality directly impacts student learning. In addressing all these gaps, this research made the following key contributions:


\begin{itemize}
    \item Comprehensive multi-metric evaluation of latest-generation LLMs for GEC, with GPT-4o establishing new state-of-the-art.
    \item Through systematic error-type analysis, we reveal that latest LLMs exhibit highly correlated correction performance for GEC ($\rho=0.947$).
    \item Our hybrid LLM-as-a-Judge framework reveals 73.76\% of corrections differing from gold standards are equally valid or preferred.
\end{itemize}

\section{Related Work}


Sequence tagging GEC approaches, like GeCTOR \cite{omelianchuk-etal-2020-gector} which use pre-trained encoders to predict specific edit operations, struggle with interconnected errors which require broader contextual understanding. Seq2Seq GEC models \cite{yuan-briscoe-2016-grammatical} which approach GEC as a monolingual translation task from incorrect to correct sentences overcame this but suffered from over-correction tendencies. Alternatively, LLMs approach GEC through conditional text generation with fine-tuned LLaMA-2 exhibiting comparative performance \cite{omelianchuk_2024}. Additionally, Omelianchuk et al. \cite{omelianchuk_2024} achieved state-of-the-art performance through edit-span level majority voting ensemble. Despite extensive comparative analysis \cite{kobayashi_2024}, latest-generation LLMs haven't been adequately compared to established GEC systems.


Reference-based metrics (ERRANT \cite{bryant_2017}) penalize valid corrections that differ from provided references, failing to account for the diversity of valid corrections, with  20–40 point improvements when references were adjusted to accept any valid correction \cite{rozovskaya_2021}. Alternatively, reference-free metrics (IMPARA \cite{Maeda_2022}) are biased toward minimal edits and struggle with scoring comprehensive corrections \cite{Bryant_2023}. Thus, LLM-as-a-Judge \cite{kobayashi-2024-b} have emerged as promising alternatives, offering inter-annotator agreement levels comparable to human evaluators at lower cost. However, relying on single LLM for evaluation risks introducing biases.

\section{Methodology}


\subsection{Multi-metric Evaluation of LLMs}
\label{subsec:multi_metric}

In this research, we evaluated three LLMs, GPT-4o \footnote{\href{https://openai.com/index/gpt-4o-system-card/}{GPT-4o System Card}}, LLaMA-3.3-70B-Instruct \cite{llama3_2024}, and DeepSeek-V3-671B \cite{deepseek_2025} in two settings: (1) Zero-shot evaluation which tests whether pre-training alone provides sufficient GEC capability, and (2) fine-tuning which tests whether task-specific adaptation can bring LLMs to competitive GEC performance levels. However, DeepSeek-V3 was excluded from fine-tuning due to the computational costs of fine-tuning a 671B-parameter model.

To capture performance across multiple dimensions we use five evaluation metrics: ERRANT F$_{0.5}$ judges whether systems make reliable corrections without introducing new errors as incorrect feedback can mislead students about grammatical rules. GLEU \cite{napoles_2016} assesses whether corrections produce naturally fluent text, important for modeling target language norms without creating over-corrected sentences. PT-ERRANT \cite{gong_2022} evaluates meaning retention, ensuring systems preserve student intent rather than rewriting their ideas. Reference-free metrics (IMPARA, Scribendi) provide additional validation, though our analysis, supported by previous studies \cite{Bryant_2023}, reveals significant limitations.


\subsection{Ensemble Analysis}
\label{subsec:ensemble_method}

To investigate whether latest LLMs exhibit complementary strengths, we developed four ensemble strategies combining outputs from fine-tuned GPT-4o, fine-tuned LLaMA-3.3, and zero-shot DeepSeek-V3. Each strategy employs majority voting as primary decision rule, with different fallback mechanisms for cases where no consensus exists: (1) Best Model: selects the highest-performing individual model's answer; (2) Meta-Model: tests whether a separate LLM (Qwen-2.5-7B-Instruct\footnote{\url{https://huggingface.co/Qwen/Qwen2.5-7B-Instruct}}) can act as an effective meta-evaluator to identify superior corrections; (3) Perplexity: tests whether fluency-based metrics (perplexity scores from base Qwen-2.5-7B\footnote{\url{https://huggingface.co/Qwen/Qwen-7B}}) can identify more natural corrections; and (4) N-gram: tests whether inter-correction similarity indicates consensus, using word-level n-gram overlap (n=3) with Jaccard similarity.

To quantify the advantage of ensembling, we conduct error-type analysis across 54 error categories, measuring both correction rates and false insertion rates. Additionally, we test heterogeneous ensembles combining our best performing model with Sequence Tagging and Seq2Seq systems to assess whether architectural diversity provides better ensemble performance.

\subsection{Hybrid LLM-as-a-Judge Framework with Human Validation}
\label{subsec:llm_judge}

To quantify how severely reference-based metrics underestimate performance, we use two LLMs, Claude 3.7 Sonnet\footnote{\href{https://assets.anthropic.com/m/785e231869ea8b3b/original/claude-3-7-sonnet-system-card.pdf}{Anthropic Claude 3.7 Sonnet System Card}} and DeepSeek-R1 \cite{deepseekr1_2025}, as primary LLM judges to assess corrections from our best-performing model. For each correction that differ from gold standard, judges categorize it into three classes: (1) gold reference preferred, (2) model correction preferred, or (3) both equally valid.

When both LLM judges reach consensus, their determination is considered final, as LLMs achieve inter-annotator agreement comparable to humans for grammatical judgments \cite{kobayashi-2024-b}. Furthermore, by requiring agreement between two, we mitigate the risk of any single model's training biases favoring certain correction styles. Only when these LLMs disagree, we invoke human judgment, where two qualified human evaluators (professional English proficiency, prior GEC annotation experience, blind to system identity, access to source sentences) apply the same three-category framework for classification. Disagreements between human annotators are resolved through discussion. 






\section{Results and Analysis}


We trained on the ABC train partition of the W\&I+LOCNESS dataset from the BEA 2019 Shared Task \cite{bryant_2019}. We tested our results on the ABCN development set (including native English texts) of BEA 2019, and on CoNLL-14 test set \cite{ng_2014}. Both datasets contain essays from ESL (English as a Second Language) learners.

\subsection{Comparative Analysis of Individual Model Performance}
\label{subsec:model_performance}

Table \ref{tab:model_performance} reveals that fine-tuned GPT-4o outperforms all individual models tested, with improvements in correction accuracy (ERRANT F$_{0.5}$), fluency (GLEU), and meaning retention (PT-ERRANT). Moreover, fine-tuning proves essential, with fine-tuned GPT-4o improving by 22.07 ERRANT F$_{0.5}$ points over zero-shot GPT-4o, making zero-shot deployment unsuitable where correction accuracy directly affects student learning. Our systems' performance was further validated in GEC Shared Task\footnote{\href{https://sites.google.com/view/nlp2025ws-langeval/task/gec?authuser=0}{NLP2025 GEC Shared Task}}, where our best performing models ranked highest in all three reference-based metrics. 

Additionally, fine-tuned GPT-4o's modest advantage over fine-tuned LLaMA demonstrates the competitiveness of open-source models, which offer greater transparency and data privacy (via local deployment without sending student data to external services), addressing regulation related privacy concerns like GDPR. Alternatively, GPT-4o offers API-based deployment with minimal infrastructure requirements which can be cost effective as well.


\begin{table}[htbp]
\caption{Existing systems' results obtained from Omelianchuk et al. \cite{omelianchuk_2024} and shared task results received from organizers. Ensemble Best 7 includes all 7 models; ``GPT-4 Rank 3'' ranks outputs from best models of each type. Heterogeneous ensemble combined our fine-tuned GPT-4o with Sequence Tagging (GECToR) and Seq2Seq (T5-11B) systems.}
\centering
\scriptsize
\definecolor{lightgrey}{rgb}{0.9,0.9,0.9}
\definecolor{darkgrey}{rgb}{0.7,0.7,0.7}
\begin{tabular}{|c|c|c|c|c|c|}
\hline
\rowcolor{darkgrey} \multicolumn{6}{|c|}{\textbf{BEA Dev Dataset}} \\
\hline
\rowcolor{darkgrey} \textbf{Model} & \textbf{ERRANT} & \textbf{GLEU} & \textbf{PT-ERRANT} & \textbf{IMPARA} & \textbf{Scribendi}\\
\hline
\multicolumn{6}{|c|}{\cellcolor{lightgrey}\textbf{Our Individual Models}} \\
\hline
Fine-Tuned GPT-4o & 0.6599 & \textbf{0.8400} & 0.7064 & 0.7768 & 0.5189\\
Fine-Tuned LLaMA 3.3 & 0.6420 & 0.8281 & 0.6842 & 0.7705 & 0.4986\\
Base DeepSeek & 0.4926 & 0.7677 & 0.5666 & 0.7754 & 0.5977\\
Base GPT-4o & 0.4592 & 0.7420 & 0.5484 & 0.7564 & \textbf{0.6843}\\
Base LLaMA 3.3 & 0.4826 & 0.7345 & 0.4973 & 0.7122 & 0.4710\\
\hline
\multicolumn{6}{|c|}{\cellcolor{lightgrey}\textbf{Our Homogeneous LLM Ensemble Systems}} \\
\hline
Majority Voting + GPT-4o Fallback & 0.6607 & 0.8392 & 0.7039 & 0.7746 & 0.5128\\
Majority Voting + Qwen Fallback & 0.6249 & 0.8312 & 0.6905 & 0.7855 & 0.5294\\
Majority Voting + Perplexity Fallback & 0.6251 & 0.8304 & 0.6879 & 0.7858 & 0.5381\\
Majority Voting + N-gram Fallback & 0.6623 & 0.8347 & 0.7122 & 0.7749 & 0.5103\\
\hline
\multicolumn{6}{|c|}{\cellcolor{lightgrey}\textbf{Heterogeneous Ensemble Systems}} \\
\hline
Majority Voting + GPT-4o Fallback & \textbf{0.6648} & 0.8360 & \textbf{0.7132} & 0.7716 & 0.4938\\
\hline
\multicolumn{6}{|c|}{\cellcolor{lightgrey}\textbf{Large Language Models}} \\
\hline
Fine-Tuned LLaMA 2 7B \cite{omelianchuk_2024} & 0.5530 & 0.7985 & 0.6157 & 0.7529 & 0.5018\\
Fine-Tuned LLaMA 2 13B \cite{omelianchuk_2024} & 0.5640 & 0.8027 & 0.6399 & 0.7554 & 0.4952\\
\hline
\multicolumn{6}{|c|}{\cellcolor{lightgrey}\textbf{Seq2Seq Models (Encoder-Decoder Transformer)}} \\
\hline
Fine-Tuned T5 11B \cite{omelianchuk_2024} & 0.5860 & 0.8231 & 0.6656 & 0.7629 & 0.4929\\
Fine-Tuned FLAN 20B \cite{omelianchuk_2024} & 0.5770 & 0.8149 & 0.6630 & 0.7582 & 0.4799\\
\hline
\multicolumn{6}{|c|}{\cellcolor{lightgrey}\textbf{Sequence Tagging Systems}} \\
\hline
GeCTOR (XLNet)  \cite{omelianchuk-etal-2020-gector} & 0.5630 & 0.7687 & 0.6248 & 0.7058 & 0.4035\\
CTC-Copy \cite{zhang_2023} & 0.5270 & 0.7714 & 0.6096 & 0.7302 & 0.4154\\
EditScorer \cite{sorokin_2022} & 0.5740 & 0.7565 & 0.6285 & 0.7072 & 0.4035\\
\hline
\multicolumn{6}{|c|}{\cellcolor{lightgrey}\textbf{Ensemble and Model Ranking Approaches}} \\
\hline
Ensemble Best 7 \cite{omelianchuk_2024} & 0.6290 & 0.7854 & 0.7040 & 0.7153 & 0.4370\\
GPT-4 Rank 3 \cite{omelianchuk_2024} & 0.5810 & 0.8270 & 0.6654 & 0.7753 & 0.5374\\
\hline
\multicolumn{6}{|c|}{\cellcolor{lightgrey}\textbf{Shared Task Competing Systems}}\\
\hline
Sugiyama et al. \cite{sugiyama_2025} & 0.4283 & 0.7603 & 0.4761 & \textbf{0.8171} & -\\
Goto et al. \cite{goto_2025} & 0.6189 & 0.7597 & 0.6483 & 0.6987 & -\\
\hline
\hline
\rowcolor{darkgrey} \multicolumn{6}{|c|}{\textbf{CoNLL-14 Dataset}} \\
\hline
\rowcolor{darkgrey} \textbf{Model} & \textbf{ERRANT} & \textbf{GLEU} & \textbf{PT-ERRANT} & \textbf{IMPARA} & \textbf{Scribendi} \\
\hline
\multicolumn{6}{|c|}{\cellcolor{lightgrey}\textbf{Our Individual Models}} \\
\hline
Fine-Tuned GPT-4o & 0.6752 & \textbf{0.7183} & 0.7390 & 0.8973 & 0.6517\\
Fine-Tuned LLaMA 3.3 & 0.6287 & 0.6969 & 0.6942 & 0.8987 & 0.5511\\
Base DeepSeek & 0.5923 & 0.6917 & 0.6651 & 0.8795 & \textbf{0.6799}\\
\hline
\multicolumn{6}{|c|}{\cellcolor{lightgrey}\textbf{Our Homogeneous LLM Ensemble Systems}} \\
\hline
Majority Voting + GPT-4o Fallback & 0.6752 & \textbf{0.7183} & 0.7390 & 0.8973 & 0.6517\\
Majority Voting + Qwen Fallback & 0.6572 & 0.7145 & 0.7180 & \textbf{0.9048} & 0.6662\\
Majority Voting + Perplexity Fallback & 0.6381 & 0.7070 & 0.7084 & 0.9033 & 0.6509\\
Majority Voting + N-gram Fallback & 0.6846 & 0.7161 & \textbf{0.7433} & 0.8904 & 0.6509\\
\hline
\multicolumn{6}{|c|}{\cellcolor{lightgrey}\textbf{Heterogeneous Ensemble Systems}} \\
\hline
Majority Voting + GPT-4o Fallback & 0.6753 & 0.7138 & 0.7370 & 0.8916 & 0.6303\\
\hline
\multicolumn{6}{|c|}{\cellcolor{lightgrey}\textbf{Existing Systems}} \\
\hline
Fine-Tuned LLaMA 2 13B \cite{omelianchuk_2024} & 0.6397 & 0.6721 & 0.6935 & 0.8236 & 0.5023\\
Fine-Tuned T5 11B \cite{omelianchuk_2024} & 0.6309 & 0.7016 & 0.6991 & 0.8883 & 0.5968\\
GeCTOR (XLNet) \cite{omelianchuk_2024} & 0.6308 & 0.6716 & 0.6709 & 0.8059 & 0.5053\\
Ensemble Best 7 \cite{omelianchuk_2024} & \textbf{0.7030} & 0.6750 & 0.7400 & 0.8086 & 0.5152\\
\hline
\end{tabular}
\normalsize
\label{tab:model_performance}
\end{table}

\subsection{Analysis of Ensembles}
\label{subsec:ensemble_analysis}


\subsubsection{Ensemble Performance Analysis}

None of our ensemble configurations significantly outperform the single fine-tuned GPT-4o model (permutation test, $p>0.05$) both on BEA-dev and CoNLL-14, demonstrating that combining multiple newest LLMs provides no measurable benefit. Furthermore, on BEA-dev, fine-tuned GPT-4o achieves performance statistically equivalent to the previous best ensemble baseline \cite{omelianchuk_2024} ($p>0.05$) and also to the heterogeneous ensemble ($p>0.05$) combining different architectural families (LLMs, Seq2Seq, Sequence Tagging). These make single fine-tuned models more suitable for educational institutions, as they achieve equivalent performance to ensemble approaches.

\subsubsection{Error-Type Analysis}

\begin{table}[t]
\caption{Correction rates and false insertion rates (\%) by error type on BEA dev (ERRANT Edit Type). M=Missing, R=Replacement, U=Unnecessary.}
\centering
\scriptsize
\definecolor{lightgrey}{rgb}{0.9,0.9,0.9}
\definecolor{darkgrey}{rgb}{0.7,0.7,0.7}
\begin{tabular}{|c|c|c|c|c|c|c|}
\hline
\rowcolor{darkgrey} 
 & \multicolumn{3}{c|}{\textbf{Correction Rates}} & 
\multicolumn{3}{c|}{\textbf{False Insertion Rates}} \\
\hline
\rowcolor{darkgrey} 
\textbf{Error} & \textbf{GPT-4o} & \textbf{Llama} & \textbf{DeepSeek} & 
\textbf{GPT-4o} & \textbf{Llama} & \textbf{DeepSeek} \\
\hline
M:ADJ & 40.0 & 40.0 & 25.0 & 52.9 & 52.9 & 61.5 \\
\hline
M:ADV & 37.9 & 20.7 & 37.9 & 52.2 & 73.9 & 61.3 \\
\hline
M:DET & 82.0 & 76.7 & 74.8 & 19.5 & 19.0 & 26.5 \\
\hline
M:PREP & 61.4 & 60.8 & 55.7 & 26.2 & 25.0 & 46.1 \\
\hline
M:PUNCT & 85.7 & 81.2 & 78.2 & 21.1 & 18.1 & 38.4 \\
\hline
R:CONJ & 28.6 & 14.3 & 14.3 & 0.0 & 50.0 & 66.7 \\
\hline
R:DET & 52.8 & 51.7 & 46.1 & 27.8 & 28.0 & 42.1 \\
\hline
R:MORPH & 73.9 & 65.3 & 69.3 & 23.7 & 27.2 & 42.1 \\
\hline
R:NOUN & 41.4 & 37.9 & 22.8 & 30.2 & 32.2 & 53.4 \\
\hline
R:NOUN:NUM & 80.6 & 73.3 & 73.3 & 24.0 & 27.1 & 35.3 \\
\hline
R:PREP & 72.2 & 69.2 & 52.0 & 20.1 & 18.9 & 26.8 \\
\hline
R:PUNCT & 75.1 & 64.8 & 34.1 & 22.0 & 23.2 & 25.8 \\
\hline
R:SPELL & 96.0 & 94.7 & 96.2 & 6.7 & 6.0 & 11.1 \\
\hline
R:VERB & 48.2 & 47.9 & 26.9 & 19.3 & 18.9 & 30.8 \\
\hline
R:VERB:FORM & 77.1 & 78.1 & 76.2 & 24.2 & 24.8 & 38.6 \\
\hline
R:VERB:SVA & 84.5 & 84.5 & 90.8 & 27.0 & 27.3 & 38.6 \\
\hline
R:VERB:TENSE & 65.7 & 63.8 & 45.8 & 22.8 & 22.9 & 21.2 \\
\hline
U:DET & 73.9 & 65.8 & 54.9 & 24.0 & 28.2 & 37.6 \\
\hline
U:PUNCT & 44.1 & 40.9 & 40.9 & 41.2 & 42.2 & 70.8 \\
\hline
\end{tabular}
\normalsize
\label{tab:rates}
\end{table}

To understand why ensembles fail, we analyzed per-error-type performance patterns across 54 error categories (Table \ref{tab:rates}). Spearman rank correlations reveal strong agreement in correction rates with fine-tuned GPT-4o and LLaMA showing near-perfect correlation ($\rho=0.947$, $p<0.0001$). This is because all models demonstrate strong performance on closed-class grammatical categories with limited lexical variation: determiners, prepositions, and rule-governed morphological transformations (subject-verb agreement, noun number, spelling). These error types rely on syntactic patterns and morphological rules well-represented in pre-training corpora and correctable through pattern matching. 

Additionally, all models struggle with open-class semantic decisions requiring contextual understanding. Correction rates remain below 50\% for content word replacements (nouns, verbs, adjectives, adverbs), reflecting that these decisions require contextual reasoning beyond pattern matching. For instance, replacing ``big'' with ``large'' depends on subtle differences and domain conventions that vary across contexts which are not encoded in distributional statistics alone.

False insertion patterns show strong correlation with $\rho$ values from 0.54 to 0.82 (all $p<0.0001$), with false insertion rates exceeding 50\% for adverbs and adjectives. This bias is inherited from pre-training on text where explicit modification is common, creating pressure toward elaboration even when original text is grammatically correct. These demonstrate that modern LLMs converge on similar error patterns due to shared architectural principles (Transformers), training objectives (next-token prediction), and training data (web text). 

\subsection{Reference-Based Metrics Underestimate Performance}
\label{subsec:reference_limitations}


\subsubsection{Limitations of Metrics}

Multiple linguistic elements can be expressed in numerous grammatically valid ways: punctuation varies across linguistic backgrounds; articles and prepositions present multiple valid options because many languages lack article systems entirely, and structure spatial and temporal relationships differently (e.g., ``at the beach'' vs. ``on the beach''); verbs offer multiple valid forms through different tense choices (e.g., ``people are suffering'' vs. ``people suffer''). This directly disadvantages multilingual learners whose diverse language choices are marked incorrect by reference-based metrics despite being grammatically valid. Alternatively, Table \ref{tab:model_performance} shows that Scribendi score would rank base GPT-4o as the best system, despite its poor correction precision (0.4592 ERRANT F$_{0.5}$). This occurs because Scribendi's Levenshtein distance penalizes extensive corrections. System optimized for reference-free metrics like this would learn to make fewer corrections, not prioritize comprehensive correction if necessary. If these metrics penalize valid corrections or overestimate performance, educators selecting GEC systems for classroom adoption, may reject systems that actually produce superior feedback for their students.

\subsubsection{Hyrbid LLM-as-a-Judge}


Among 1,730 corrections where fine-tuned GPT-4o produced different edits to gold references, two LLM judges reached consensus in 64.34\% of cases, preferring our model's corrections over gold standards in 30.87\% of cases, with 13.70\% corrections rated equally valid ($\kappa = 0.72$, 82\% overall agreement of LLM with human experts). Following human evaluation to resolve all LLM disagreements (inter-annotator agreement, Cohen's $\kappa$: 0.975), final results revealed GPT-4o's corrections were preferred for 35.61\% of edits, gold standards for 26.24\%, while 38.15\% were judged equally valid. So, in 73.76\% of cases where our model produced corrections which differed from gold standards, these were judged as either superior or equally valid. Finally, this approach reduces traditionally time-intensive human evaluation effort by 64\%, making rigorous correction quality assessment more feasible.



\section{Conclusion}

We demonstrate that fine-tuned GPT-4o achieves high correction precision with strong fluency and meaning retention, establishing performance reliable enough for immediate student feedback. However, our error-type analysis reveals LLMs as GEC assistants excel at closed-class grammatical patterns while struggling with open-class semantic decisions, making ensembling less effective and benefits learners from languages without article systems (e.g., Mandarin, Japanese) more compared to learners from Romance languages who make fewer article errors. Additionally, our hybrid LLM-as-a-Judge framework reveals that reference-based metrics underestimate GEC performance with 73.76\% of corrections, diverging from gold standards, judged equally valid or even superior. Despite these findings, some limitations apply as LLMs may have encountered test datasets during pre-training, potentially inflating performance.


%
%
%
\bibliographystyle{splncs04}
\bibliography{custom2, custom, anthology}

\end{document}